\documentclass[runningheads]{llncs}
\usepackage{graphicx}
\usepackage{amssymb}
\usepackage{graphicx}
\usepackage{amssymb}
\usepackage[margin=1.45in]{geometry}
\usepackage{multirow}
\usepackage{subcaption}
\usepackage{color,soul}
\usepackage{authblk}
\usepackage{dblfnote}
\usepackage{hyperref}

\begin{document}

%
%

\author{Tran Thi Hong Hanh\inst{1,2,4} \orcidID{0000-0002-5993-1630} \and
Antoine Doucet\inst{2}\orcidID{0000-0001-6160-3356} \and
Nicolas Sidere\inst{2} \orcidID{0000-0001-6719-5007}\and
Jose G. Moreno\inst{3}\orcidID{0000-0002-8852-5797} \and
Senja Pollak\inst{4}\orcidID{0000-0002-4380-0863}}

\authorrunning{Tran et al.}

\institute{University of Science and Technology in Hanoi, Vietnam \and
University of La Rochelle, L3i
laboratory, France \and
University of Toulouse - IRIT, France \and
Jozef Stefan Institute,  Ljubljana, Slovenia}

\title{Named entity recognition architecture combining contextual and global features}

\maketitle

\fbox{\begin{minipage}{36.5em}
\noindent The  final reviewed  publication  was published in ICADL proceedings as part of the Lecture Notes in Computer Science book series (LNCS, volume 13133) and part of the Information Systems and Applications, incl. Internet/Web, and HCI book sub series (LNISA, volume 13133). The publication is available online at \href{https://doi.org/10.1007/978-3-030-91669-5_21}{https://doi.org/10.1007/978-3-030-91669-5\_21}.
\end{minipage}}

\begin{abstract}
Named entity recognition (NER) is an information extraction technique that aims to locate and classify named entities (e.g., organizations, locations,...) within a document into predefined categories. Correctly identifying these phrases plays a significant role in simplifying information access.
However, it remains a difficult task because named entities (NEs) have multiple forms and they are context dependent. While the context can be represented by contextual features, the global relations are often misrepresented by those models. In this paper, we propose the combination of contextual features from XLNet and global features from Graph Convolution Network (GCN) to enhance NER performance. Experiments over a widely-used dataset, CoNLL 2003, show the benefits of our strategy, with results competitive with the state of the art (SOTA). 

\end{abstract}
\vspace{-0.2cm}
\section{Introduction}

The proliferation of large digital libraries has spurred interest in efficient and effective solutions to manage the collections of digital contents (documents, images, videos, etc.) which are available, but not always easy to find. As an alternative to better handle information in digital libraries, named entity recognition (NER) was introduced.


NER is an information extraction technique that aims to locate named entities (NEs) in text and classify them into predefined categories. Correctly identifying entities plays an important role in natural language understanding and numerous applications such as entity linking, question answering, or machine translation, to mention a few.


A crucial component that contributes to the recent success of NER progress is how meaningful information can be captured from original data via the word embeddings, which can be divided into two major types: global features and contextual features (in the scope of this paper, ``features'' and ``embeddings'' are interchangeable terms).
\begin{itemize}
    \item \textbf{Global features} \cite{pennington2014glove} capture latent syntactic and semantic similarities. They are first constructed from a global vocabulary (or dictionary) of unique words in the documents. Then, similar representations are learnt based on how frequently the words appear close to each other. The problem of such features is that the words' meaning in varied contexts is often ignored. That means, given a word, its embedding always stays the same in whichever sentence it occurs. Due to this characteristic, we can also define global features as ``\textit{static}''. Some examples are word2vec \cite{church2017word2vec}, GloVe \cite{pennington2014glove}, and FastText \cite{joulin2016fasttext}.

    \item \textbf{Contextual features} \cite{devlin2019bert} capture word semantics in context to address the polysemous and context-dependent nature of words. By passing the entire sentence to the pretrained model, we assign each word a representation based on its context, then capture the uses of words across different contexts. Thus, given a word, the contextual features are ``\textit{dynamically}'' generated instead of being static as the global one. Some examples are ELMo \cite{peters2018deep}, BERT \cite{devlin2019bert}, and XLNet \cite{yang2019xlnet}. 
\end{itemize}

In terms of global features, there exist several tokens that are always parts of an entity. 
The most obvious cases, as an example in the CoNLL 2003 dataset,  are the names of countries include U.S. (377 mentions), Germany (143 mentions), Australia (136 mentions), to mention a few. However, it is not true for all tokens in an entity. The token may or may not be part of an entity (e.g, ``Jobs said'' vs. ``Jobs are hard to find'') and may belong to different entity types depending on the context (e.g, ``Washington'' can be classified as a person or a location). Meanwhile, the contextual features are based on neighboring tokens, as well as the token itself. They aim to represent word semantics in context to solve the problem of using global features, so as to improve the prediction performance (e.g, ``Jobs'' in ``Jobs said'' and ``Jobs are hard to find'' will have different representations).

In this paper, we present a joint architecture to enhance the NER performance simultaneously with static and dynamic embeddings \footnote{Link to the code: github.com/honghanhh/ner-combining-contextual-and-global-features}. Extensive experiments on CoNLL 2003 dataset suggest that our strategy surpasses the systems with standalone feature representation. The main contributions of this paper are: 
\begin{itemize}
    \item We introduce a new architecture that combines the contextual features from XLNet and the global features from GCN to enhance NER performance.
    \item We demonstrate that our model outperforms the systems using only contextual or global features alone and has a competitive result compared with SOTAs on CoNLL 2003 dataset.
\end{itemize}

This paper is organised as follows: Section~\ref{sec:sota} presents the related work, which leads to our approach's descriptions in Section~\ref{sec:method} and the corresponding experimental details in Section~\ref{sec:expe}. The results are reported in Section~\ref{sec:results}, before we conclude and present future works in Section~\ref{sec:conclusion}.
\vspace{-0.2cm}
\section{Related work}
\label{sec:sota}
\subsection{Named entity recognition}
The term “named entity” (NE) first appeared in the $6^{th}$ Message Understanding Conference (MUC-6) \cite{grishman1996message} to define the recognition of the information units. 
Regarding the surveys on NER techniques \cite{palshikar2013techniques,yadav2019survey,li2020survey}, we can broadly divide them into four categories: Rule-based, unsupervised learning, feature-based supervised learning, and deep learning based approaches.

\noindent \textbf{Rule-based approaches}
Rule-based NER is the most traditional technique that does not require annotated data as it relies on  manually-crafted rules well-designed by the domain experts (e.g., LTG \cite{mikheev1999named}, NetOwl \cite{krupka2005description}). Despite good performance when the lexicon is exhaustive, such systems often achieve high precision and low recall due to the limitation on domain-specific rules and incomplete dictionaries.

\noindent \textbf{Unsupervised learning}
Another approach that also needs no annotated data is unsupervised learning, typically NE clustering \cite{collins1999unsupervised}. The key idea is to extract NEs from the clustered groups based on context similarity. The lexical resources, lexical patterns, and statistics are computed on a large corpus and then applied to infer mentions of NEs. Several works proposed the unsupervised systems to extract NEs in diverse domains \cite{etzioni2005unsupervised,nadeau2006unsupervised}.

\noindent \textbf{Feature-based Supervised learning}
Given annotated data, features are carefully designed so that the model can learn to recognize similar patterns from unseen data. 
Several statistical methods have been proposed, notably Markov models, Conditional Random Fields (CRFs), and Support Vector Machines (SVMs). Among them, CRF-based NER has been widely applied to identify entities from texts in various domains 
\cite{liu2020hamner,ritter2011named,rocktaschel2012chemspot}.
However, these approaches depend heavily on hand-crafted features and domain-specific resources, which results in the difficulty to adapt to new tasks or to transfer to new domains.

\noindent \textbf{Deep Learning}
Neural networks offer non-linear transformation so that the models can learn complex features and discover useful representations as well as underlying factors. Neural architectures for NER often make use of either Recurrent  Neural  Networks (RNNs)  or  Convolution  Neural  Networks (CNNs) in conjunction with  CRFs \cite{chiu2016named} to  extract information automatically. With further researches on contextual features, RNNs plus LSTM units and  CRFs have been proposed \cite{lample2016neural} to improve the performance.  Moreover, the conjunction of bidirectional LSTMs, CNNs, and CRFs \cite{ma2016end} is introduced to exploit both word- and character-level representations. The combination of Transformer-based models, LSTMs, and CRFs \cite{liu2018empower} is also applied to extract knowledge from raw texts and empower the NER performance. 
\vspace{-0.3cm}
\subsection{Embeddings}
A key factor that contributes to the success of NER is how we capture meaningful information from original data via word representations, especially global features and contextual features.

\noindent \textbf{Global features}
Global features are context-free word representations that can capture meaningful semantic and syntactic information. It can be represented at different levels such as word-level features \cite{liao2009simple}, lookup features \cite{hoffart2011robust}, document and corpus features \cite{ji2016joint}. 
Recently, the global sentence-level representation \cite{zhang-etal-2018-sentence} has been proposed to capture global features more precisely and it outperforms various sequence labeling tasks. Furthermore, the Graph Neural Network \cite{yao2019graph} is getting more attention to not only have rich relational structure but also preserve global structure information of a graph in graph embeddings. 

\noindent\textbf{Contextual features}
Contextual features are context-aware word representations that can capture word semantics under diverse linguistic contexts. That is, a word can be represented differently and dynamically under particular circumstances. The contextual embeddings are often pretrained on large-scale unlabelled corpora and can be divided into 2 types: unsupervised approaches \cite{lample2019cross,lan2019albert} and supervised approaches  \cite{subramanian2018learning}. 

The contextual embeddings succeed in exploring and exploiting the polysemous and context-dependent nature of words, thereby moving beyond global word features and contributing significant improvements in NER. In contrast, the global features are still less-represented. 
\vspace{-0.2cm}
\section{Methodology}
\label{sec:method}

In this section, we explain how we extract global as well as contextual features and how to combine them. For global features, we take advantage of GCN \cite{seti2020named,cetoli2017graph} to better capture the correlation between NEs and the global semantic information in text, and to avoid the loss of detailed information. For contextual features, we apply XLNet \cite{yang2019xlnet}, a Transformer-XL pretrained language model that exhibits excellent performance for language tasks by learning from bi-directional context.
The details are explained in the following subsections.

\subsection{GCN as Global Embeddings}

Graph Convolutional Network (GCN) aims to learn a function of signals/features on a graph $G=(V,E)$ with $V$ as Vertices and $E$ as Edges. Given $N$ as number of nodes, $D$ as number of input features, and $F$ as the number of output features per node, GCN takes 2 inputs:
(1) An $N \times D$ feature matrix $X$ as feature description; (2) An adjacency matrix $A$ as representative description of the graph; Finally, it returns as output $Z$, an $N \times F$ feature matrix \cite{duvenaud2015convolutional}.

Every neural network layer can then be written in the form of a non-linear function:
\begin{equation}
    H^{(l+1)} = f(H^{(l)},A)
\end{equation}
where $H^{(0)} = X$, $H^{(L)} = Z$,  L being the number of layers. 

In our specific task, we capture the global features by feeding feature matrix X and adjacent matrix A into a graph using two-layer spectral convolutions in GCN. Raw texts are first transformed into word embeddings using GloVe. Then, universal dependencies are employed so that the input embeddings are converted into graph embeddings where words become nodes and dependencies become edges. After that, two-layer GCN is applied to the generated matrix of nodes feature vectors X and the adjacent matrix A to extract meaningful global features.

Mathematically, given a specific graph-based neural network model $f(X, A)$, spectral GCN follows the layer-wise propagation rule:
\begin{equation}
\label{gcn_h}
H^{(l+1)} = \sigma(\tilde{D}^{\frac{-1}{2}}\tilde{A}\tilde{D}^{\frac{-1}{2}}H^{(l)}W^{(l)})
\end{equation}
where $A$ is the adjacency matrix, $X$ is the matrix of node feature vectors (given sequence x), $D$ is the degree matrix, $f(\cdot)$ is the neural network like differentiable function, $\tilde{A} = A + I_{N}$ is the adjacency matrix of the undirected graph G with added self-connections,  $I_{N}$ is the identity matrix of N nodes, $\tilde{D}_{i} = \sum_j \tilde{A}_{ij}$, $W^{(l)}$ is the layer-specific trainable weight matrix, $\sigma(\cdot)$ is the activation function, and $H^{(l)} \in \mathbb{R}^{(N \times D)}$ is the matrix of activation in the $l^{th}$ layer (representation of the $l^{th}$ layer), $H^{(0)} = X$.

After calculating the normalized adjacency matrix $\tilde{D}^{\frac{-1}{2}}\tilde{A}\tilde{D}^{\frac{-1}{2}}$ in the preprocessing step, the forward model can be expressed as:
\begin{equation}
    Z = f(X,A) = softmax(\tilde{A}ReLU(\tilde{A}XW^{0})W^{1})
\end{equation}
where $W^{(0)} \in R^{C \times H}$ is the input-to-hidden weight matrix for a hidden layer with H feature maps and $W^{(1)} \in R^{H \times F}$ is the hidden-to-output weight matrix.

$W^{(0)}$ and $W^{(1)}$ are trained using gradient descent. The weights before feeding into Linear layer with Softmax activation function are taken as global features to feed into our combined model. We keep the prediction results of GCN after feeding weights to the last Linear layer to compare the performance and prediction qualities with our proposed architecture's results.

\subsection{XLNet as Contextual Embeddings}
XLNet is an autoregressive pretraining method based on a novel generalized permutation language modeling objective. Employing Transformer-XL as the backbone model, XLNet exhibits excellent performance for language tasks involving long context by learning from bi-directional context and avoiding the disadvantages in the autoencoding language model.

The contextual features are captured from the sequence using permutation language modeling objective and two-stream self-attention architecture, integrating relative positional encoding scheme and the segment recurrence mechanism from Transformer-XL \cite{yang2019xlnet}. Given a sequence x of length T, the permutation language modeling objective can be defined as: 

\begin{equation}
\max _{\theta} \mathbb{E}_{\mathbf{z} \sim \mathcal{Z}_{T}}\left[\sum_{t=1}^{T} \log p_{\theta}\left(x_{z_{t}} \mid \mathbf{x}_{\mathbf{z}_{<t}}\right)\right]
\end{equation}
where $\mathcal{Z}_{T}$ is the set of all possible permutations of the index sequence of length T 
$[1, 2, . . . , T]$, $z_{t}$ is the $t^{th}$ element of a permutation $\mathbf{z} \in \mathcal{Z}_{T}$,  $\mathbf{z}<t$ is the first $(t - 1)^{th}$ elements of a permutation $\mathbf{z} \in \mathcal{Z}_{T}$, and  $p_{\theta}$ is the likelihood. $\theta$ is the parameter shared across all factorization orders during training so $x_{t}$ is able to see all $x_{i} \neq x_{t}$ possible elements in the sequence.

We also use two-stream self-attention to remove the ambiguity in target predictions. For each self-attention layer  $m = 1, ... , M$, the two streams of representation are updated schematically with a shared set of parameters:
\begin{equation}
\label{eqhztm}
\begin{array}{l}
g_{z_{t}}^{(m)} \leftarrow Attention \left(\mathrm{Q}=g_{z_{t}}^{(m-1)}, \mathrm{KV}=\mathbf{h}_{\mathrm{z}<t}^{(m-1)} ; \theta\right) \\
h_{z_{t}}^{(m)} \leftarrow Attention \left(\mathrm{Q}=h_{z_{t}}^{(m-1)}, \mathrm{KV}=\mathbf{h}_{\mathrm{z} \leq t}^{(m-1)} ; \theta\right)
\end{array}
\end{equation}

where  $g_{z_{t}}^{(m)}$ is the query stream that uses $z_t$ but cannot see $x_{z_t}$, $h_{z_{t}}^{(m)}$ is the content stream that uses both $z_t$ and $x_{z_t}$, and K, Q, V are the key, query, value, respectively.

To avoid slow convergence,  the objective is customized to maximize the log-likelihood of the target sub-sequence conditioned on the non-target sub-sequence as in Equation \ref{eqmaxtheta}.

{\small
\begin{equation}
\label{eqmaxtheta}
\max _{\theta} \mathbb{E}_{\mathbf{z} \sim \mathcal{Z}_{T}}\left[ \log p_{\theta}\left(x_{\mathbf{z}_{> c}} \mid \mathbf{x}_{\mathbf{z}_{\leq c}}\right)\right]    =   \mathbb{E}_{\mathbf{z} \sim \mathcal{Z}_{T}}\left[\sum_{t = c , the+ 1}^{\mid z \mid} \log p_{\theta}\left(x_{z_{t}} \mid \mathbf{x}_{\mathbf{z}<t}\right)\right]
\end{equation}
}
where $\mathbf{z}_{> c}$ is the target sub-sequence, $\mathbf{z}_{\leq c}$ is the non-target one, and c is the cutting point.

Furthermore, we make use of the relative positional encoding scheme and the segment recurrence mechanism from Transformer-XL. 
While the position encoding ensures the reflection in the positional information of text sequences, the attention mask is applied so the texts are given different attention during the creation of input embedding. 
Given 2 segments $\mathbf{x} = s_{1:T}$ and $\mathbf{x} = s_{T:2T}$ from a long sequence s,
$\mathbf{z}$ and z referring to the permutations of [1 ... T] and [T + 1 ... 2T], we process the first segment, and then cache the obtained content representations $\mathbf{h}^{(m)}$ for each layer m. After that, we update the attention for the next segment x with memory, which can be expressed as in Equation \ref{eqhztm}.
\begin{equation}
\label{eqhztm}
h_{z_{t}}^{(m)} \leftarrow  Attention\left(\mathrm{Q}=h_{z_{t}}^{(m-1)}, \mathrm{KV}=\left[\tilde{\mathbf{h}}^{(m-1)}, \mathbf{h}_{\mathrm{z} \leq t}^{(m-1)}\right] ; \theta\right)
\end{equation}

Similar to global features, we capture the weights before feeding to the last Linear layer and use it as contextual embeddings of our combined model. For the purpose of comparison, we also keep the prediction results of XLNet after feeding weights to the last Linear layer.
\vspace{-0.3cm}
\subsection{Joint architecture}
\begin{figure}[ht]
  \centering
  \includegraphics[width=10cm]{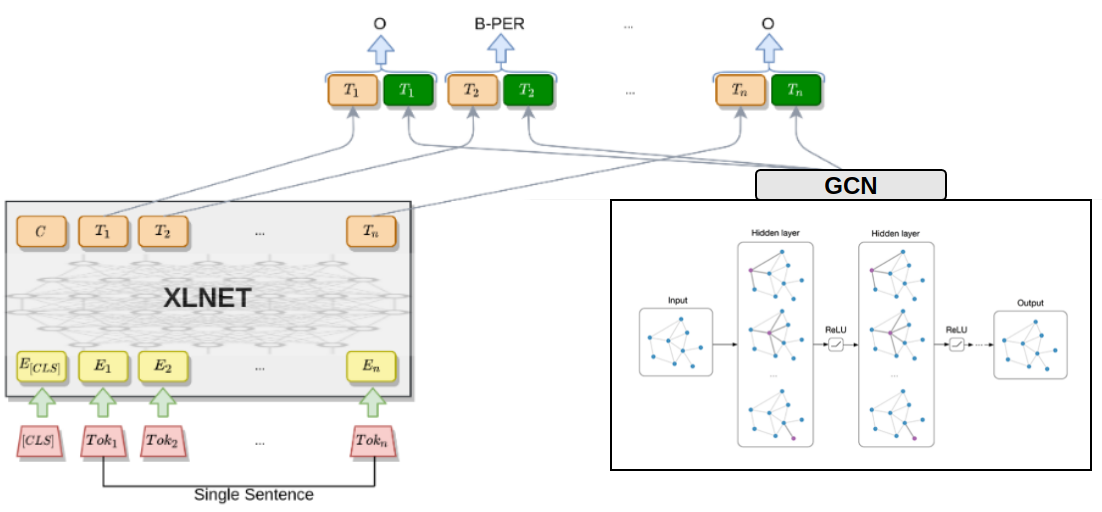}
  \vspace{-0.3cm}
  \caption{Visualization of the global architecture of our proposed approach.}
  \label{fig:mix}
\end{figure}

Given global and contextual features from GCN and XLNet, respectively, we concatenate and feed them into a Linear layer, which is simplest way to show the most evident impact of these features to the NER task. The proposed approach is presented in Fig. \ref{fig:mix}. 
\vspace{-0.2cm}
\section{Experimental setup}
\label{sec:expe}

In this section, we describe the dataset, the evaluation metrics, as well as present our implementations and experimental configurations on XLNet, GCN, and the joint models in detail. 
\vspace{-0.7cm}
\subsection{Dataset}
We opted for the CoNLL 2003 \cite{tjongkimsang2003conll}, one of the widely-adopted benchmark datasets for NER tasks. The English version is collected from the Reuters Corpus with news stories between August 1996 and August 1997. The dataset concentrates on 4 types of NEs: persons (PER), locations (LOC), organizations (ORG), and miscellaneous (MISC). 
%
\vspace{-0.3cm}
\subsection{Implementation details}
\noindent\textbf{Global embeddings with GCN}
The sentences are annotated with universal dependencies from spaCy to create a graph of relations 
where words become nodes and dependencies become edges. The dataset is then converted into 124 nodes and 44 edges with the training corpus size of approximately 2 billion words, the vocabulary size of 222,496, and the dependency context vocabulary size of 1,253,524. Next, the graph embeddings are fed into 2 Graph Convolution layers with a Dropout of 0.5 after each layer to avoid overfitting. The global features are captured before the last Linear layer. We perform batch gradient descent using the whole dataset for every training iteration, which is a feasible option as long as the dataset fits in memory. We take advantage of TensorFlow for efficient GPU-based implementation of Equation \ref{gcn_h} using sparse-dense matrix multiplications. 

\noindent\textbf{Contextual embeddings with XLNet}
We have investigated on diverse embeddings such as FastText \cite{mikolov2018advances} \footnote{https://fasttext.cc/}, Flair \cite{akbik2019flair} \footnote{https://github.com/flairNLP/flair}, Stanza \cite{qi2020stanza} \footnote{https://github.com/stanfordnlp/stanza} and XLNet \cite{yang2019xlnet} \footnote{https://github.com/zihangdai/xlnet} pretrained embeddings. Preliminary results suggest that XLNet (XLNet-Base, Cased) outperforms others, therefore, is chosen for our final implementation. 
The word embedding of size 768 with 12 layers were used for XLNet. Each layer consists of 3 sublayers: XLNet Relative Attention, XLNet Feed Forward, and Dropout layer. The XLNet Relative Attention is a two-stream self-attention mechanism as mentioned in Equation \ref{eqhztm}. A Normalization layer with element-wise affine and a Dropout layer are employed around this sub-layer. Meanwhile, XLNet Feed Forward is a fully connected feed-forward network, whose outputs are also of dimension 768, the same as the outputs of the embedding layers. Like the previous sublayers, the Feed Forward layer is surrounded by a Normalization layer and a Dropout layer, however, another 2 Linear layers are added between them. Then, an additional Dropout layer is counted. It is notable that we only take the rate of 0.1 for every Dropout layer inside our model, from sublayers to inside sublayers.  After 12 XLNet layers, another Dropout layer is added before the last Linear layer. We capture the intermediary output before the last Linear layer as the contextual features. 

\noindent\textbf{Proposed Model}
Additional steps were taken to maintain alignments between input tokens and their corresponding labels as well as to match corresponding representations from global features to contextual features in the same sentence. First, we define an attention mask in XLNet as a sequence of 1s and 0s, with 1s for the first sub-word as the whole word embedding after tokenization and 0s for all padding sub-words. Then, in GCN features, we map the corresponding word representation at the position that the XLNet attention mark returns 1s and pad 0 otherwise. Therefore, each sentence has the same vector dimension in both global and contextual embeddings, which simplifies the concatenation. 

In our implementation, we used a GPU 2070 Super and a TitanX GPU with 56 CPUs, 128 GB RAM. The hyperparameters were $300$ as embedding size, $16$ as batch size,  $5$e-$5$ as learning rate, $0.5$ as dropout rate, $4$ for number of epochs.

\subsection{Metrics}




We choose ``relaxed'' micro averaged $F_1$-score, which regards a prediction as the correct one as long as a part of NE is correctly identified. 
This evaluation metric has been used in several related publications, journals, and papers on NER \cite{takeuchi2002use} \cite{huang2015bidirectional} \cite{lample2016neural} \cite{ma2016end}.
\vspace{-0.2cm}
\section{Results}
\label{sec:results}
We conducted multiple experiments to investigate the impact of global and contextual features on NER.  Specifically, we implemented the architecture with only global features, only contextual features, and then the proposed joint architecture combining both features. 

As shown in Table \ref{table:joint_res}a, the proposed model achieves \textbf{93.82\%} in $F_1$-score, which outperforms the two variants using global or contextual features alone. In terms of recognition of specific entity types, the details are provided in Table \ref{table:joint_res}b, showing that PER is the category where the best results are achieved, while the lowest results are with the MISC, that is, the category of all NEs that do not belong to any of the three other predefined categories. Note that using only training data and publicly available word embeddings (GloVe), our proposed model has competitive results without the need of adding any extra complex encoder-decoder layers. 
\begin{table}[htb]
    \caption{Evaluation on the prediction results of our proposed model.}
   \begin{subtable}[t]{.5\textwidth}
        \raggedleft
        \centering
        \caption{Results of the proposed joint architecture compared to only contextual or only global features.}
        \begin{tabular}{lc} 
        \hline
        \textbf{Embeddings}                 & \textbf{$F_{1}$ scores}   \\ 
        \hline
        Global features                  &       88.63             \\
        Contextual features              &       93.28             \\ \hline
        Global + contextual features     &       \textbf{93.82}      \\ \hline
        \end{tabular}
    \end{subtable}
    \begin{subtable}[t]{.52\textwidth}
        \caption{Performance evaluation per entity type.}
        \vskip 5pt
        \raggedright
        \centering
        \begin{tabular}{ccccc}
        \hline
                            \textbf{Entity types} & \textbf{Precision} & \textbf{Recall} & \textbf{$F_1$-score} &  \\ \hline
        \multicolumn{1}{c}{\textbf{LOC}}  &   94.15    &    93.53      &      93.83            \\ 
        \multicolumn{1}{c}{\textbf{MISC}} &   81.33    &    81.89     &       81.62          \\ 
        \multicolumn{1}{c}{\textbf{ORG}}  &   88.97    &    92.29      &      90.60           \\ 
        \multicolumn{1}{c}{\textbf{PER}}  &   96.67    &    97.09      &      96.88         \\ \hline
        
        \end{tabular}
    \end{subtable}%
\label{table:joint_res}
\end{table}

\begin{figure}[ht]
  \centering
  \includegraphics[width=8cm]{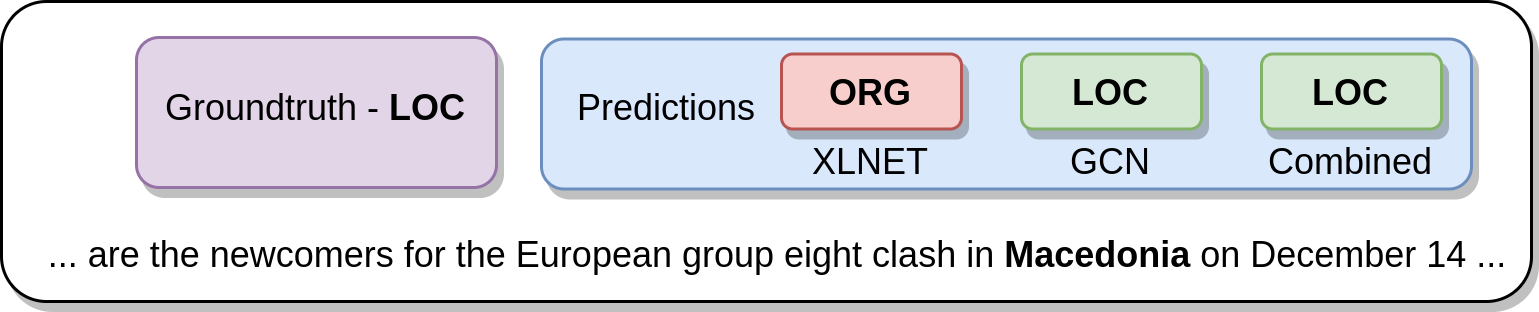}
  \vspace{-0.3cm}
  \caption{XLNet, GCN, and the combined model's prediction on CoNLL 2003's example.}
  \label{fig:comp}
\end{figure}

Furthermore, the benefit of the joint architecture is illustrated in Fig. \ref{fig:comp}. While contextual features (XLNet), which is used in the majority of recent SOTA approaches, misclassifies the entity, the prediction from GCN and the combined model correctly tags ``MACEDONIA'' as the name of a location, confirming our hypothesis on the effect of global features.


\begin{figure}[ht]
  \centering
  \includegraphics[width=10cm]{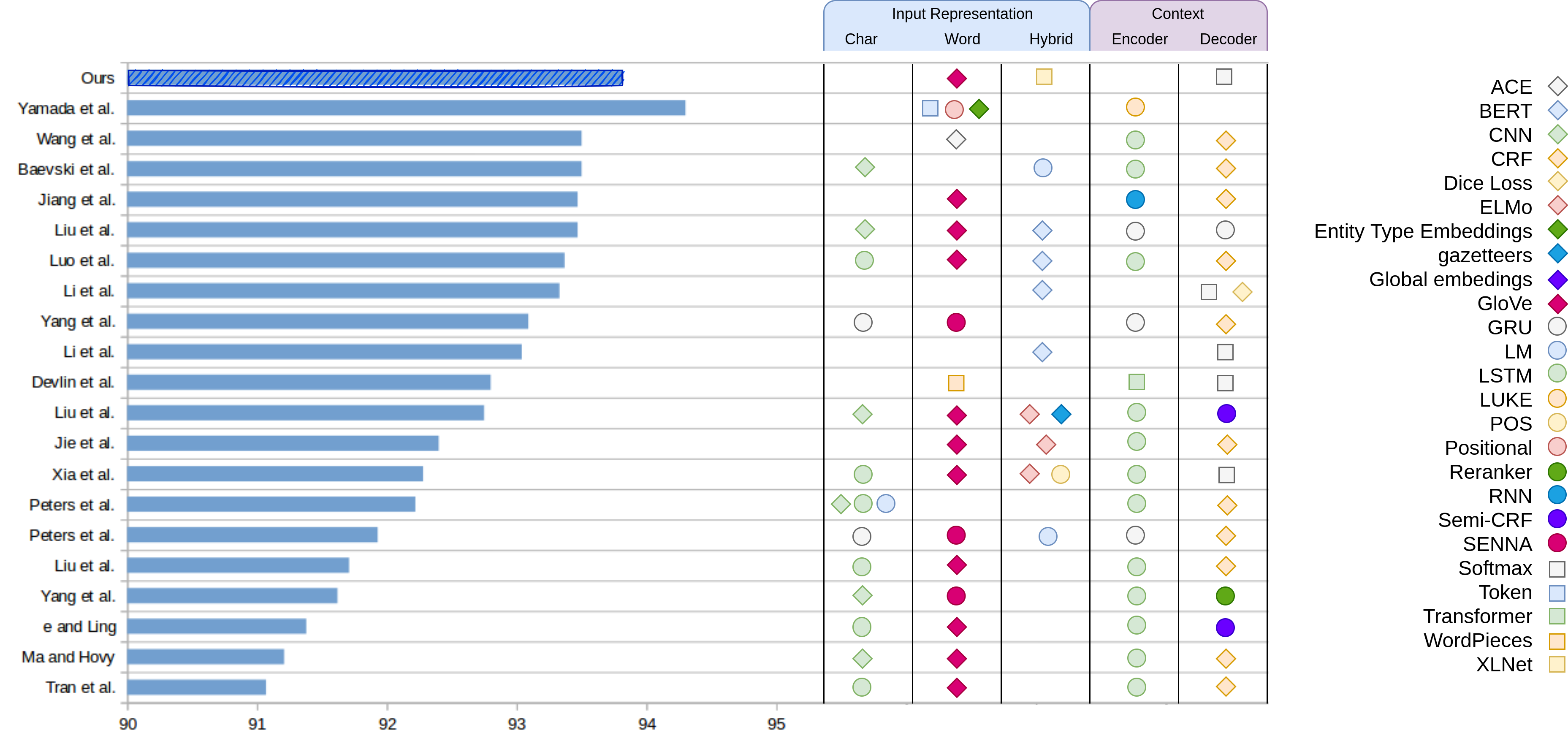}
    \vspace{-0.3cm}
  \caption{Comparison of our proposal against SOTA techniques on the CoNLL 2003 dataset in terms of $F_1$-score. Values were taken from original papers and sorted by descending order.}
  \label{fig:method-comparison}
\end{figure}

In Figure \ref{fig:method-comparison}, we compare our results with reported SOTA results on the same dataset from 2017 up to now. It can be observed that our results are competitive compared with SOTA approaches as the difference is by a small margin (the current benchmark is 94.3 \% $F_1$-score, compared to 93.82 \%  achieved by our approach). Moreover, we notice that NER performance can be boosted with external knowledge (i.e. leveraging pretrained embeddings), as proven in our approach as well as in top benchmarks \cite{liu2019towards,liu2019gcdt,luo2020hierarchical}. More importantly, complex decoder layers (CRF, Semi-CRF,...) do not always lead to better performance in comparison with softmax classification when we take advantage of contextualized language model embeddings.

\vspace{-0.3cm}
\section{Conclusion and future work}
\label{sec:conclusion}
We propose a novel hierarchical neural model for NER that uses both global features captured via graph representation and contextual features at the sentence level via XLNet pretrained model. The combination of global and contextual embeddings is proven to have a significant effect on the performance of NER tasks. Empirical studies on the CoNLL 2003 English dataset suggest that our approach outperforms systems using only global or contextual features, and is competitive with SOTA methods. Given the promising results in English, our future work will consist of adapting the method to other languages, as well to a cross-lingual experimental setting. In addition, we will consider further developing the method by also incorporating background knowledge from knowledge graphs and ontologies.

\subsection*{Acknowledgements}
This work has been supported by the European Union's Horizon 2020 research and innovation program under grants 770299 (NewsEye) and 825153 (EMBEDDIA). The work of S. P. has also received financial support from the Slovenian Research Agency for research core funding for the Knowledge Technologies programme  (No. P2-0103) and the project CANDAS (No. J6-2581).

%

\bibliographystyle{splncs04}
\bibliography{ref}

\end{document}